\documentclass{article}

\usepackage{tkai}

\usepackage[utf8]{inputenc} 
\usepackage[T1]{fontenc}    
\usepackage{hyperref}       
\usepackage{url}            
\usepackage{booktabs}       
\usepackage{amsfonts}       
\usepackage{nicefrac}       
\usepackage{microtype}      
\usepackage[numbers]{natbib}
\usepackage{booktabs} 
\usepackage{graphicx,subcaption,float}
\usepackage{amsmath}
\usepackage{mathtools}
\usepackage[toc,page]{appendix}
\usepackage{enumitem}
\usepackage{graphics}
\usepackage{graphicx}
\usepackage{algorithm}
\usepackage{algorithmicx}
\usepackage[noend]{algpseudocode}
\algnewcommand{\LineComment}[1]{\State \(//\) #1}
\usepackage{enumitem}
\usepackage{xr}

\usepackage{etoolbox}
\usepackage{tikz}
\usetikzlibrary{tikzmark}
\usetikzlibrary{calc}
\usepackage{pgfplots}

\errorcontextlines\maxdimen

\newcommand{\ALGtikzmarkcolor}{black}
\newcommand{\ALGtikzmarkextraindent}{4pt}
\newcommand{\ALGtikzmarkverticaloffsetstart}{-.5ex}
\newcommand{\ALGtikzmarkverticaloffsetend}{-.5ex}

\makeatletter
\newcommand*{\addFileDependency}[1]{
  \typeout{(#1)}
  \@addtofilelist{#1}
  \IfFileExists{#1}{}{\typeout{No file #1.}}
}
\makeatother

\makeatletter
\newcounter{ALG@tikzmark@tempcnta}

\newcommand\ALG@tikzmark@start{%
    \global\let\ALG@tikzmark@last\ALG@tikzmark@starttext%
    \expandafter\edef\csname ALG@tikzmark@\theALG@nested\endcsname{\theALG@tikzmark@tempcnta}%
    \tikzmark{ALG@tikzmark@start@\csname ALG@tikzmark@\theALG@nested\endcsname}%
    \addtocounter{ALG@tikzmark@tempcnta}{1}%
}

\def\ALG@tikzmark@starttext{start}
\newcommand\ALG@tikzmark@end{%
    \ifx\ALG@tikzmark@last\ALG@tikzmark@starttext
    \else
        \tikzmark{ALG@tikzmark@end@\csname ALG@tikzmark@\theALG@nested\endcsname}%
        \tikz[overlay,remember picture] \draw[\ALGtikzmarkcolor] let \p{S}=($(pic cs:ALG@tikzmark@start@\csname ALG@tikzmark@\theALG@nested\endcsname)+(\ALGtikzmarkextraindent,\ALGtikzmarkverticaloffsetstart)$), \p{E}=($(pic cs:ALG@tikzmark@end@\csname ALG@tikzmark@\theALG@nested\endcsname)+(\ALGtikzmarkextraindent,\ALGtikzmarkverticaloffsetend)$) in (\x{S},\y{S})--(\x{S},\y{E});%
    \fi
    \gdef\ALG@tikzmark@last{end}%
}

\apptocmd{\ALG@beginblock}{\ALG@tikzmark@start}{}{\errmessage{failed to patch}}
\pretocmd{\ALG@endblock}{\ALG@tikzmark@end}{}{\errmessage{failed to patch}}
\makeatother

\title{Like a bilingual baby:\\The advantage of visually grounding a bilingual language model}

\author{Khai-Nguyen Nguyen \\
  Bucknell Univeristy\\
  Lewisburg, PA \\
  \texttt{nkn002@bucknell.edu} \\\And
  Zixin Tang \\
  The Pennsylvania State University \\
  State College, PA \\
  \texttt{zqt5035@psu.edu} \\\And
  Ankur Mali \\
  University of South Florida \\
  Tampa, FL \\
  \texttt{ankurarjunmali@usf.edu} \\\And
  M. Alex Kelly \\
  Carleton University \\
  Ottawa, Canada \\
  \texttt{alex.kelly@carleton.ca} \\}
\pgfplotsset{compat=1.17}

\begin{document}
\setlength{\abovedisplayskip}{0.065cm}
\setlength{\belowdisplayskip}{0pt}

\maketitle

\begin{abstract}
Unlike most neural language models, humans learn languages in a rich, multi-sensory, and, often, multi-lingual environment. Conversely, language models are typically trained on only language data, and often on only one language. We hypothesize that perceptual grounding scaffolds language learning, including learning relationships between languages. To better understand multilingualism and the role of visual input in language understanding, we train a recurrent language model on images and corresponding text in English and Spanish from MS-COCO-ES. We find that visual grounding improves the model's understanding of semantic similarity within and across languages and improves language generation. Our results provide evidence of the advantages of visually grounded language models. We posit that language learning is better understood as integral to the totality of a learner's experiences, and thus there is a need for more naturalistic language data from multilingual speakers and multilingual datasets with perceptual grounding.

\end{abstract}

\section{Introduction}

With the effects of globalization on business, education, and culture, multilingualism---speaking two or more languages---is becoming more and more common. The prevalence of multilingualism and speech patterns peculiar to multilingualism creates a need for computational language models that can handle, process, and generate the language of multilingual communities. 

It is well known that a person's knowledge is inseparable from the physical or social context in which it is learned and used; as humans, we create a world model based on context \cite{greeno1993situativity}. Perceptual symbols theory states that language, reasoning, context, and cognition are grounded in perceptual features that provide visual clues to create world models \cite{barsalou1999perceptions}. 
Unlike state-of-the-art language models, humans learn languages in a rich perceptual environment. Perceptual data contributes to linguistic tasks and plays an important role in the acquisition of language in humans \cite{barsalou2008grounded, goodman2007bayesian}. Perceptual grounding facilitates first language acquisition (e.g., the illustrations in children’s picture books) and second language acquisition (e.g., studying abroad). By using perceptual information, we can quickly understand the meaning of a new word when learning a language by mapping it to its real-world subject. 
Language models that incorporate visual data have a stronger correspondence to human judgments of word similarity and human reaction times on semantic priming tasks, especially for concrete nouns and visually descriptive words \cite{bruni2014multimodal,kiela2017learning,kievit2011semantic,merkz2022}.

Recently it has been shown that recurrent neural network-based models are equivalent to universal computational models such as Turing Machines, even with finite precision \cite{chung2021turing, mali2019neural, stogin2020provably} and even with bounded time \cite{stogin2020provably}. As the brain is widely understood to be a kind of Turing machine \cite{speer2016ensemble, mcculloch1943logical}, RNNs are an obvious choice to conduct this study. On the other hand, the self-attention layers of the popular transformer models \cite{vaswani2017transformer} have restricted capability and fail to recognize context-free languages, even when allowed infinite precision in the weights \cite{hahn2020theoretical}, and have issues generalizing to unseen distributions. Hence to better study the role of visual grounding in multilingual language learning, we extend a Long Short Term Memory (LSTM) recurrent model with multimodal and multilingual inputs, trained on images and corresponding English and Spanish text. Our interest in images is largely due to the availability of visual datasets rather than a commitment to the vision to the exclusion of other senses being important for human-like language learning. Congenitally blind people’s language experience is perceptually grounded, just not visually grounded.

We aim to understand the process of using perceptual information in human language learning in language models, specifically facilitating multilingual learning using visual information, to examine if visual representation allows the model to understand better the relationship between words from different languages with the same semantic meaning. Perceptually grounded multilingual language models have the potential to be \textbf{(1)} more human-like in how they process multiple languages (i.e., better models of multilingual speakers) and \textbf{(2)} scaffold the acquisition of multiple languages given conditions of plentiful perceptual data and limited linguistic data (as is generally  the case for human second language learners). 

In what follows, we discuss related work on perceptually grounded and multilingual language modeling, describe  our model, and present results (namely, overall perplexity in Spanish and English and both within and between language judgements of semantic similarity). We find that the use of visual information lowers perplexity and improves correlation to human judgements of semantic similarity both between and within languages. However, the performance improvement is least-significant for abstract words. Our results align with prior studies on images and monolingual data \cite{ororbia2018like, merkz2022}; visual grounding improves multilingual model performance on next-word prediction and semantic alignment across languages. 

\section{Related Work}

The importance of visual or perceptual features to language comprehension is widely studied in neuro-imaging, and behavioral studies \cite{barsalou1999perceptions, barsalou2008grounded}. These studies provide substantial evidence that language and perception can benefit each other. This can also be seen in large vision-language models, where pre-training on language benefits visual processing \cite{radford2021learning, huo2021wenlan}. It is also evident from studies showing an infant's world model is efficiently created by jointly learning different modalities \cite{frank2008bayesian}. Children or infants rapidly learn new words by inferring and analyzing information from their physical world \cite{alishahi2008fast}. Bayesian cognitive models have captured the rapid dynamics of children's language acquisition by pairing syntactic information from language experience with semantic knowledge from world experience, such the learning in the two modalities bootstrap off of each other \cite{abend2017bootstrapping}. Thus, integrating vision and language can help us better understand language acquisition in human brains and can benefit artificial intelligence systems through efficient learning and reasoning.

Multilingual language models succeed in many tasks, including language comprehension, generation, cross-lingual translations, and information retrieval. Studies have found that after fine-tuning on target language pairs, 
pre-trained models can successfully handle multiple multilingual tasks, including but not limited to next-word prediction, translation, language generation for code-switching sentences, and speech recognition \cite{lee2020balm, gupta2020semi, gao2019ganbert}. 
 However, achieving human-like performance on tasks involving integration and interactions among multiple languages is still challenging. \cite{winata2021metaembedding} found that even though pre-trained models succeed in multiple multilingual tasks, they may not perform well in forming representations of code-switching patterns in language production, indicating a lack of sufficiently deep integration and interactions between models' representations across languages. 
 
 Furthermore, questions of how multilingual models integrate knowledge across languages and if that integration is human-like remain a matter of study \cite{Putnam2017integrated}. Studies have found that even when multilingual models form language-integrated representations \cite{dhar2018does} and align similar knowledge among languages under similar representations \cite{chi2020UGBERT}, the models still may not achieve satisfying results on higher-level multilingual tasks such as translation alignment and language generation. 
 In short, more studies are needed to understand better the mechanisms of language representations of multilingualism and how representations of different languages are involved in the language generation process.

\section{Model Design}

Using multilingual data, we will evaluate gated recurrent models such as long-short-term memory (LSTM). In this study, we show that having vision context or semantic input helps the model better understand the syntactic structure across languages.
We first define the vanilla LSTM architecture, which is defined as follows:

\begin{align}
    \mathbf{i_t} = \sigma(\mathbf{W}^i x_t + \mathbf{V}^i \mathbf{z_{t-1}}) \\
    \mathbf{f_t} = \sigma(\mathbf{W}^f x_t + \mathbf{V}^f \mathbf{z_{t-1}}) \\
    \mathbf{g_t} = \phi(\mathbf{W}^g x_t + \mathbf{V}^g h_{t-1} ) \\
     \mathbf{o_t} = \sigma(\mathbf{W}^o x_t +  \mathbf{V}^o \mathbf{z_{t-1}}) \\
    \mathbf{c_t} = \mathbf{f_t} \odot \mathbf{c_{t-1}} + \mathbf{i_t} \odot \mathbf{g_t} \\
    \mathbf{z_t} = \mathbf{o_t} \odot \mathbf{\phi(c_t)} 
\end{align}

\noindent where $\mathbf{W} \in \mathbb{R}^{m \times n}$ are input to hidden synaptic weights with $m$ rows and $n$ columns, $\odot$ represents the Hadamard product, $\mathbf{V} \in \mathbb{R}^{n \times m}$ are hidden to hidden to hidden synaptic weights, $\sigma(x)$ is the sigmoid ($e^x/(1+e^x)$) activation function, $\phi(x)$ is the hyperbolic tangent ($2\sigma(2x)-1)$, $\mathbf{z}_t$ represents the hidden layer at time $t$, $\mathbf{i}_t$ represents the input gate at time $t$, $\mathbf{f}_t$ represents the forget gate at time $t$, $\mathbf{o}_t$ represents the output gate at time $t$ and $\mathbf{c}_t$ represents the cell state of the network.
To integrate visual context information into the LSTM, we fuse or augment the network with a computer vision system, motivated by the prior work focused on image captioning \cite{xu2015show}. We fine-tune one of the state-of-the-art vision model (Resnet-50) \cite{resnet50} on our benchmarks. The vision model was originally designed to perform classification, but by training it on a large corpus, it develops semantic representations. We extract semantic-level features by extracting synaptic weights, the visual context $c$, from the final pooling layer, which we use to provide the language model with visually grounded information. This visual information is then incorporated into the LSTM by multiplying it with a learnable synaptic connection matrix $\mathbf{M}$ and altering the model's hidden state $\mathbf{z_t}$ at time $t$ as follows: 
\begin{equation}\label{eqn:matrixM}
    \mathbf{z_t} = [\mathbf{o_t} \odot \phi(\mathbf{c_t})] \odot (\mathbf{Mc}).
\end{equation}
where $\mathbf{c_t}$ is the cell state at time $t$, $\mathbf{o_t}$ is the output gate, and $\odot$ is the Hadamard product. 

\paragraph{Incorporating multilingualism}
We utilize the multilingual pre-trained BPEmb subword embedding \cite{heinzerling2018bpemb} containing $320,000$ words and sub-words from $275$ languages creating an embedding of size $N \times 320000$. This allows the model to be trained and tested in a multilingual setting. This corresponds to initializing $W_i$ in the LSTM.

\paragraph{Unimodal Multilingual LSTM}
The unimodal multilingual LSTM (UM-LSTM) is an LSTM language model that utilizes the multilingual BPEmb embedding for multilingual learning. The unimodal multilingual LSTM architecture serves as the \textbf{baseline model} in our experiments.

\paragraph{Multimodal Multilingual LSTM}

The multimodal multilingual LSTM (MM-LSTM), based on the multimodal LSTM \cite{ororbia2018like}, but without peephole connection and utilizes the multilingual BPEmb embedding for multilingual learning. The MM-LSTM model takes two streams of input: the language input stream and the visual input stream. In the language input stream, the input text is tokenized and processed by the multilingual embedding layer and becomes the input to the LSTM. The input images are processed in the visual input stream by a frozen pre-trained ResNet50 \cite{resnet50}. We extract the vector produced from ResNet50's final pooling layer to obtain a distributed representational vector of the image based on equation \ref{eqn:matrixM}.

\section{Experiments}
\paragraph{MS-COCO-ES} The MS-COCO-ES dataset\footnote{\url{https://github.com/carlosGarciaHe/MS-COCO-ES}} contains  100,000 human-annotated English captions from the MS-COCO dataset with around 19K unique tokens and 100,000 Spanish captions machine-translated from the English captions with around 21K unique tokens. Each image has five English captions and five Spanish captions. We split this data into training/validation/test sets with a ratio of 80/10/10.

\noindent 
\subsection{Training}
We train both models, UM-LSTM and MM-LSTM, on the MS-COCO-ES training set with a batch size of $32$ and a sequence length of $32$ tokens for $15$ epochs. 
Unlike the next-step prediction model proposed in prior work \cite{ororbia2018like}, all our models utilize the sequence-to-sequence training setup: Given an input sequence from $0$ to $t-1$, we predict its output from $1$ to $t$. We use the cross-entropy or negative log-likelihood loss function. As noted earlier, this work is focused on understanding how visual features contribute to multi-lingual language understanding. To do so, all our models are optimized to minimize negative-log likelihood. Our models focus on language understanding instead of specific tasks undertaken by image captioning or machine translation models. In other words, these models rank the output distribution based on the plausibility of the candidate output. We also clip the gradients to $2$ to avoid vanishing/exploding gradient issues. We use stochastic gradient descent with an initial learning rate $\lambda = 1.0$ that is halved based on validation performance using patience scheduling $=3$. Both models have one LSTM layer followed by a dropout layer with a dropout rate of $0.2$. \footnote{The hyperparameters are chosen based on prior work \cite{ororbia2018like}.} We train the models on an NVIDIA GeForce RTX $2080$ Ti with $12$ GB of RAM. We evaluate the models' ability to understand the relationship between pairs of words in monolingual and cross-lingual contexts using semantic similarity judgment. All experiments are conducted for $3$ trials with different random seeds. We report mean performance and standard error for all models; this also helps evaluate the uncertainty in our model's prediction. 

\noindent 
\subsection{Semantic Similarity Judgement }
The similarity of a pair of words can be calculated as the cosine similarity of the corresponding embedding vectors of the pair. We can evaluate the correctness of a model by comparing to human judgements of word similarity, using data gathered by explicitly asking participants to evaluate the synonymy or category of words \cite{baroni2014don,bruni2014multimodal, derby2020analysing, kiela2017learning, Landauer1997, speer2016ensemble} or tacitly inferred from semantic priming response times \cite{Jones2006,merkz2022}. 

\subsubsection{Data}  
The linguistic definition of word \textit{relatedness} (e.g., \textit{coffee} and \textit{cup}) and word \textit{similarity} (e.g., \textit{coffee} and \textit{tea}) may differ from the understanding that experiment participants have when asked to explicitly rate the similarity or relatedness of pairs of words \cite{hill2015simlex}. If the difference between relatedness and similarity is not specified to participants, the nature of the ratings is ambiguous. In our experiment, both relatedness and similarity are used as a metric to evaluate the models' semantic understanding \cite{merkz2022}. We consider four relatedness and/or similarity datasets and their derivatives in Table \ref{tab:r_squared}.

\begin{itemize}
    \item \textbf{SimLex-999} \cite{hill2015simlex} consists of 999 pairs of words in English scored on a scale of 0 to 10 specifically by semantic similarity rather than relatedness.
    \item \textbf{MEN} \cite{bruni2014multimodal} consists of 3000 pairs of words in English scored by semantic relatedness on a scale of 0 to 50. Data was collected by showing two word pairs at a time to subjects and asking them to choose the more related pair. Words in MEN have strong visual references due to the use of visual words.
    \item \textbf{WordSim353} \cite{finkelstein2001placing} consists of 353 pairs of words in English scored by semantic similarity on a scale of 0 to 10. Since the participants are not instructed to judge for relatedness or similarity, the score is considered hybrid by recent studies. \cite{agirre2009study} split the dataset into two subsets, one for evaluating similarity (WordSim353-S) and one for evaluating relatedness (WordSim353-R). We abbreviate the WordSim datasets as WS.
    \item \textbf{RG-65} \cite{rubenstein1965contextual} consists of 65 pairs of words in English scored by semantic similarity on a scale of 0 to 4. Pairs are scored by examining the proportion of words common to the contexts of words A and B.  \cite{camacho2015framework} presents a Spanish version of the dataset (RG-$65_{ES}$) with the same number of word pairs as the original and a cross-lingual version in English and Spanish of the dataset (RG-$65_{EN-ES}$) with 126 word pairs.
\end{itemize}

\begin{table}[!htb]
\caption{Number of word pairs originally in each dataset and number of word pairs used for evaluation. Pairs are dropped if they have at least one word not in the embedding matrix}
\begin{center}
\begin{tabular}{lcc}
\hline
{\textbf{Dataset}} & Original & Used\\
\hline
SimLex-999  & 999 & 731\\
MEN & 3000 & 2037\\
WordSim-S & 203 & 160\\
WordSim-R & 252 & 195\\
WordSim353 & 353 & 274\\
RG-$65_{EN-ES}$ & 126 & 53\\
RG-$65_{ES}$ & 65 & 18\\
RG-65 & 65 & 41\\
\hline
\end{tabular}

\label{tab:used_data}
\end{center}
\end{table}
Following the procedures in \cite{merkz2022}, for each model, we calculate the cosine similarity of the output embedding vectors of each word pair. We then compute the $r$ correlation coefficient between the cosine and the human annotations for each model. Pairs with at least one word not in the output embedding matrix are dropped from the calculation. The number of used pairs can be found at Table~\ref{tab:used_data}. 
We also calculate the partial correlation of the MM-LSTM model with the UM-LSTM model as the control variable.

\begin{table*}
\caption{The Pearson correlation between the two models and human annotations, the partial-$r$ of the MM-LSTM with the UM-LSTM as the control variable, and the adjusted $p$-value using the procedure in \cite{benjamini1995controlling} ($\ast: p <.05, \ast\ast: p < .01$) for each LSTM hidden unit. A greater partial-$r$ means a greater portion of correlation in the human annotations is explained by the MM-LSTM than the baseline UM-LSTM model. Note: UM is Unimodal Multilingual model and MM stands for Multimodal Multilingual model.}
\small
\begin{center}
\begin{tabular}{lcccccccc}
\hline
{\textbf{\#Hidden units}} & SimLex-999 & MEN & WS-S & WS-R & WS353 & RG-$65_{{EN-ES}}$ & RG-$65_{{ES}}$ & RG-65
\\
\hline
\hline
\textbf{\#Hidden units=128} \\
$r_{UM-LSTM}$ & -0.055 & 0.327 & 0.293 & 0.024 & 0.154 & 0.372 & 0.317 & 0.352\\
$r_{MM-LSTM}$ & \textbf{-0.018} & \textbf{0.425} & \textbf{0.326} & \textbf{0.03} & \textbf{0.176} & \textbf{0.486} & \textbf{0.426} & \textbf{0.457}\\
\textit{partial-}$r$ & 0.096 & 0.314 & 0.156 & 0.09 & 0.068 & 0.336 & 0.547 & 0.333\\
\textit{p-value} &* & ** & & & & * & * &\\
\hline
\textbf{\#Hidden units=256}\\
$r_{UM-LSTM}$ & -0.046 & 0.364 & 0.337 & 0.043 & 0.183 & 0.421 & 0.392 & 0.373\\
$r_{MM-LSTM}$ & \textbf{-0.013} &\textbf{ 0.477} & \textbf{0.358} & 0.030 & \textbf{0.190} & \textbf{0.523} & \textbf{0.493} & \textbf{0.472}\\
\textit{partial-}$r$ & 0.094 & 0.369 & 0.127 & -0.028 & 0.052 & 0.343 & 0.435 & 0.35\\
\textit{p-value} &* & ** & & & & * & & *\\
\hline

\textbf{\#Hidden units=512}\\
$r_{UM-LSTM}$ & -0.052 & 0.386 & 0.342 & 0.046 & 0.188 & 0.533 & 0.436 & 0.472\\
$r_{MM-LSTM}$ & \textbf{-0.01} & \textbf{0.493} & \textbf{0.356} & \textbf{0.058} & \textbf{0.199} & \textbf{0.571} & \textbf{0.531} & \textbf{0.536}\\
\textit{partial-}$r$ & 0.132 & 0.365 & 0.105 & 0.042 & 0.068 & 0.284 & 0.431 & 0.290\\
\textit{p-value} & ** & ** & & & & & & \\
\hline


\hline

\end{tabular}

\label{tab:r_squared}
\end{center}
\end{table*}

\begin{table}[ht]
\caption{The test perplexity (Lower is better \textcolor{green}{$\downarrow$}) of the UM-LSTM (baseline) and MM-LSTM averaged over $3$ trials for each LSTM's hidden unit size. VL+VL represents the model trained and tested with visual information, whereas VL+L indicates the model trained using visual features but tested without a visual cue. Note: CotM represents the model with fixed or non-trainable memory matrix $\mathbf{M}$.}
\small
\centering
\begin{tabular}{rcccc}
\hline
\textbf{Model} & \textbf{\textit{n}=128} & \textbf{\textit{n}=256} & \textbf{\textit{n}=512}\\
\hline
\verb|UM-LSTM| & 18.20 & {15.29}& 14.30 \\
\verb|MM-LSTM|(\textbf{ours}) &  & & \\

\verb|VL+VL|(\textbf{ours}) &{\textbf{15.0} $\pm$ 0.1} \textcolor{green}{$\downarrow$} &{\textbf{13.28} $\pm$ 0.08}\textcolor{green}{$\downarrow$} & {\textbf{13.18} $\pm$ 0.15}\textcolor{green}{$\downarrow$}  \\
\verb|VL+L|({ours}) &{41.5 $\pm$ 1.6}\textcolor{red}{$\uparrow$}  & {42.04 $\pm$ 1.07}\textcolor{red}{$\uparrow$}& {46.4 $\pm$ 2.48}\textcolor{red}{$\uparrow$}\\
\verb|CotM| &  & & \\
\verb|VL+VL| &20.55 & 17.26&16.26 \\
\verb|VL+L| &20.63 & 17.35&16.14 \\
\hline
\end{tabular}

\label{fig:losses}
\end{table}

\section{Results}

As hypothesized, visually grounded information better assists the MM-LSTM model in learning English and Spanish. Table \ref{fig:losses} illustrates how the MM-LSTM consistently outperforms the baseline UM-LSTM model and shows better generalization on the MS-COCO-ES dataset. In particular, to show the consistency of our results across various settings, we conduct experiments with varied hidden layer sizes ($n = 128, 256$ or $512 $). It can be seen throughout all hidden layer sizes that models augmented with visual clues during testing and training consistently outperform language-only models by a wide margin. Whereas it is interesting to see when we remove visual clues while testing (VL+L), the model has difficulty understanding the language. This correlates with psychological studies where authors have shown that the absence of partial vision hampers language understanding \cite{wilson2008loss, kolarik2017partial}. Hence our ``blinded" model (VL+L) shows losing visual capability hampers the overall performance. It is interesting to note that the vision+language model learns different vector space embedding representations compared to the language-only model. This is  evident from the VL+L model performance, where removing visual or semantic cues leads to poor performance. To further validate our hypothesis, we fixed the learnable memory synaptic matrices $\mathbf{M}$. In other words, the memory matrix is randomly initialized and never learned during training, and thus the model does not learn to depend on it to represent linguistic information.  
As evident from our result in Table \ref{fig:losses}, a model with non-learnable memory matrices ($\mathbf{M}$), referred to as CotM in our table, behaves nearly identically in the presence or absence of visual cues at test, whereas the model that jointly learns our memory matrix achieves enhanced language understanding as long as it still has access to visual information when tested. This result supports  the importance of jointly learning vision and language.

\paragraph{Similarity-based Evaluation} We next focus on a more complex scenario concerning similarity-based evaluation. As hypothesized in this work, visual features are more human-like and can better facilitate human language learning and comprehension. Thus we evaluate our model performance with human-annotated word similarity scores. In particular, we report the Pearson correlation between human similarity scores and the models' similarity scores in Table \ref{tab:r_squared}. As evident from our experiments, both models positively correlate with a human-annotated score, thus showing both approaches learn useful information. However, the model coupled with visual clues consistently outperforms the language-only model and thus is much closer to human evaluation. This indicates that the MM-LSTM is more human-like than the UM-LSTM. Looking at the partial-R values in all but the WordSim-R, WordSim-S, and WordSim353 datasets, we see that the MM-LSTM's similarity scores explain a significant correlation to the human-annotated scores with the baseline model as the control variable. 
While both models get negative correlation scores in the SimLex-999 dataset, partial-R values still hold for the SimLex-999 dataset, despite having negatively correlated human-annotated scores. One reason for having a negative correlation but positive R-values can be attributed to credit assignment issues which plague the generalization capability of LSTM-based models \cite{ororbia2020continual, mali2021investigating}, resulting in sub-optimal performance on a few scenarios. Future work on designing forward propagation and local learning approaches might provide  evidence of the importance of visual features in more difficult datasets like WordSim.
\noindent\begin{table}[t]
\begin{tabular}[]{l l}%
\hline\\
\includegraphics[width=0.2\columnwidth, height=0.2\columnwidth]{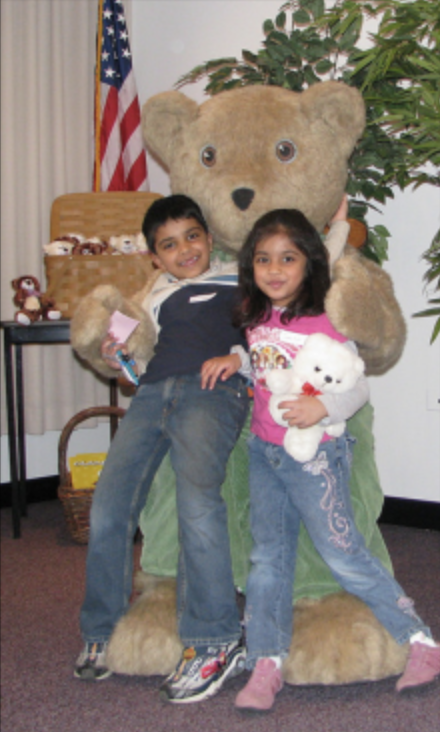}&
\begin{minipage}{0.7\columnwidth}%
- dos niños con osos de peluche sentados en una silla con un oso\\
- dos niños con osos de peluche sentados en una silla con una muñ\\
-a large brown and brown teddy bear standing in front of a stuffed animal in the

\end{minipage}
\\
\hline\\
\includegraphics[width=0.2\columnwidth, ]{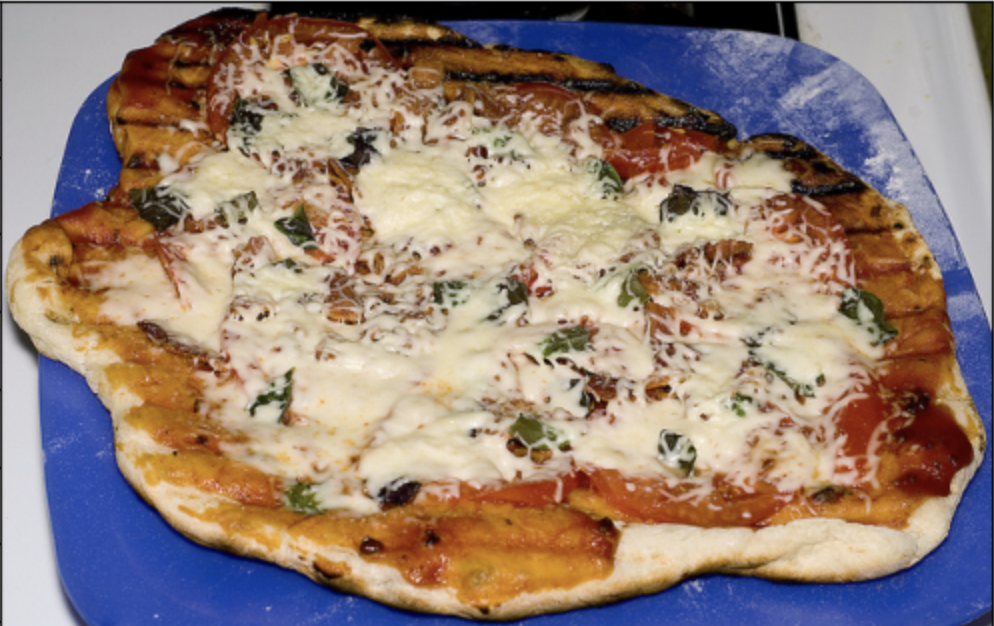}&
\begin{minipage}{0.7\columnwidth} 
- there is a pizza with cheese and tomato on it\\
- there is a pizza and pepperoni on top of\\
- un grupo de pizzas diferentes en él en una mesa\\
\end{minipage}
\\
\hline\\
\includegraphics[width=0.2\columnwidth]{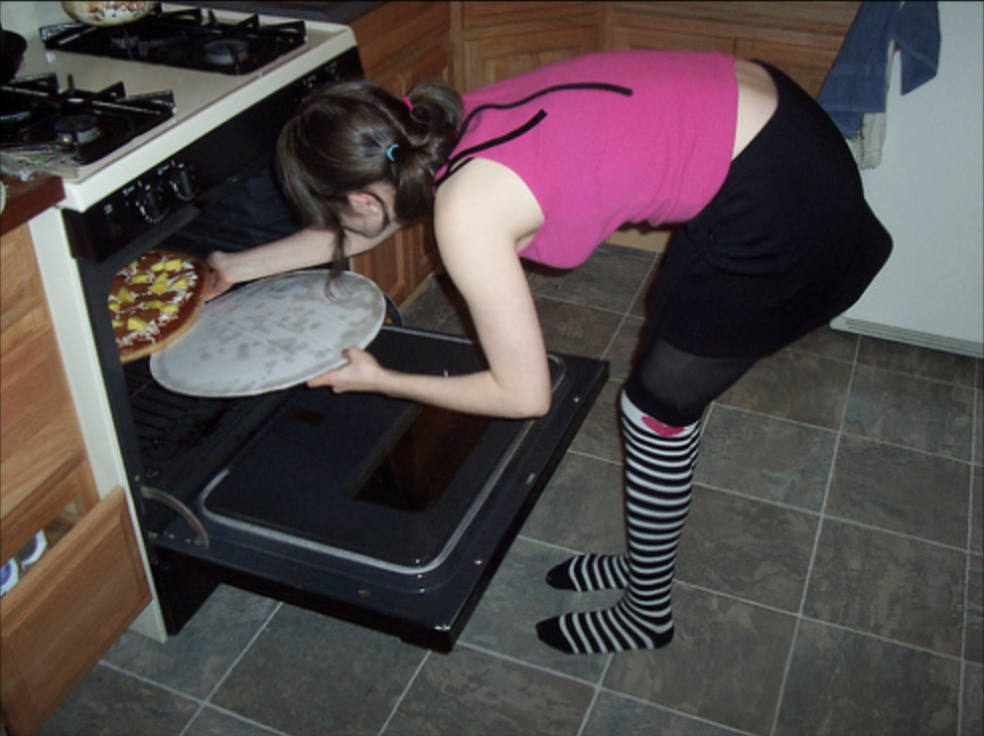}&
\begin{minipage}{0.7\columnwidth} 
- girl in a kitchen with a pizza in the kitchen in\\
- un grupo de personas en una cocina con una estufa en\\
\end{minipage}
\\
\hline\\
\includegraphics[width=0.2\columnwidth, height=0.2\columnwidth]{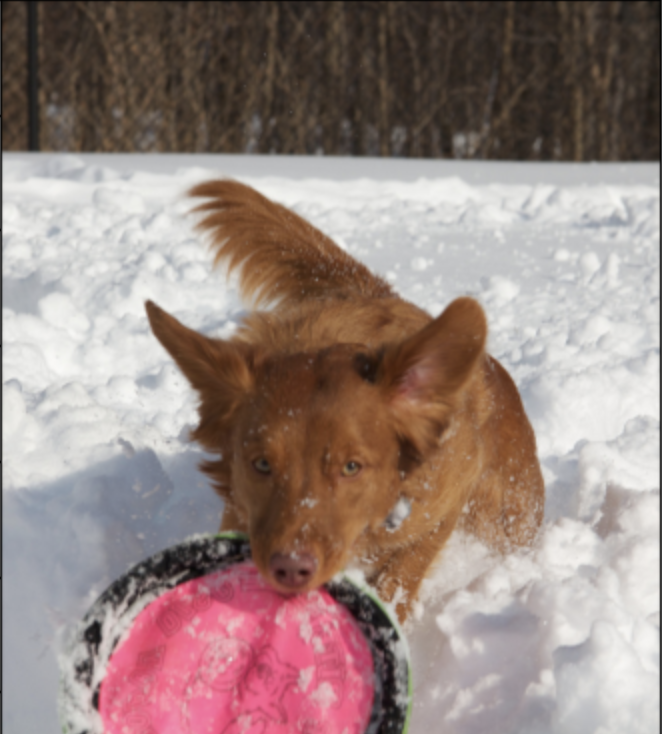}&
\begin{minipage}{0.7\columnwidth}
- a dog is laying in the snow with a frisbee in\\
- un gran grupo de perros en la nieve y un perro marrón en la\\

\end{minipage}
\end{tabular}
\caption{Generated captions by MM-LSTM on images randomly sampled from the MS-COCO-ES test set. For the prompt, we feed the first word of the true caption as the prompt for text generation in one language. To force the model to generate caption in the other language, we use the prompt "a" for English and "un" for Spanish.}
\label{tab:captions}
\end{table}

\subsection{Sampling}
As a ``sanity check" that the MM-LSTM is working correctly and can generate text in both languages, in Table \ref{tab:captions}, we include sample text generated by the MM-LSTM using image prompts. The text was generated using beam search with top-k ($k=10$), and top-p ($p=0.3$) sampling. We restrict the sampling procedure to be equal to ground-truth caption length. Finally, words are ranked based on model probabilities to generate the text.  
As evident in Table \ref{tab:captions}, the proposed model can generate text in English and Spanish, thus demonstrating both language competencies.

\section{Discussion}
To test the hypothesis that visual grounding yields better language generation and more human-like semantic similarity judgments in the context of multilingual language models, we train a multimodal multilingual Long Short-Term Memory (LSTM) neural network on images and texts in English and Spanish. We compare our proposed model performance to a unimodal multilingual LSTM trained only on texts. 
When evaluated on semantic similarity and relatedness, the multi-modal model (MM-LSTM) outperforms the baseline unimodal model (UM-LSTM) on multiple datasets in both English and Spanish. As expected, the embeddings of the MM-LSTM model are significantly more correlated than the UM-LSTM to human judgments in the MEN dataset (English words scored on relatedness) since textual information in the data shares strong visual references. Notably, MM-LSTM is significantly more correlated to human judgments of cross-lingual word similarity than the UM-LSTM on the RG-65$_{EN-ES}$ dataset, which suggests that perceptual grounding may indeed help integrate language knowledge across different languages, as we hypothesized. However, where the correlation between visual and textual information is minimal, the advantage of the visually-grounded model over the language model is unreliable and non-significant, as in the WordSim datasets. 

Our experiments on semantic similarities align with our hypothesis and show that our model acquires a better representation to facilitate improved bilingual and cross-lingual language modeling whenever the corpus supports a correlation between visual and textual context.  Thus visual information is useful for improving the performance of language models in terms of both language generation and correlation to human judgments of semantic similarity.

The study conducted in this work offers a promising research direction and is not confined to a single domain, such as machine learning. By providing further evidence for the importance of perceptual grounding, this study informs the study of language in psychology and cognitive science, and can aid in designing computational models to study language acquisition in the visually impaired or help us better understand how language acquisition is facilitated by knowledge from the physical world.


\section{Conclusion}
We explore the effect of visual grounding on a multilingual language model. We hypothesize that, like in humans, visual information can play a crucial role in the process of language learning, representation, and generation in neural language models. We propose the MM-LSTM multimodal multilingual LSTM model to study visually-aided language generation. We train the model on MS-COCO-ES, an English and Spanish image captioning dataset. In Table~\ref{fig:losses}, we observe that adding visual information during training improves language generation. In Table~\ref{tab:r_squared}, we observe that the visually grounded model better understands the relationship between words (especially nouns) within and between languages than the baseline model.
It should be noted that the role of the visual context is crucial as without it, the model performed notably worse than the baseline across our experiments. In future work, we intend to design a pre-training strategy \cite{wang2022image} that jointly learns vision-language space and allows the model to extrapolate to language generation in the absence of a visual signal.

 Our results provide preliminary evidence of the advantages of visually grounded language models. Future work in this area demands better datasets. The fact that the MS-COCO-ES dataset is small and machine-translated limits our ability to make strong claims. Currently, most image captioning datasets are unilingual, most multilingual datasets lack perceptual grounding of any kind, and most multilingual datasets are parallel corpora where languages are never mixed organically, lacking features characteristic of communication between multi-lingual speakers, such as code-switching. We posit that language learning is better understood as integral to the totality of a learner's experiences across sensory modalities, and the languages are known to the learner. Thus there is a need for more naturalistic language data from multilingual speakers and multilingual datasets with perceptual grounding.

\bibliography{zmain}

@article{vaswani2017transformer,
  title={Attention is all you need},
  author={Vaswani, Ashish and Shazeer, Noam and Parmar, Niki and Uszkoreit, Jakob and Jones, Llion and Gomez, Aidan N and Kaiser, {\L}ukasz and Polosukhin, Illia},
  journal={Advances in neural information processing systems},
  volume={30},
  year={2017}
}

@article{Jones2006,
  author = {Jones, M. N. and Kintsch, W. and Mewhort, D. J. K.},
  journal = {Journal of Memory and Language},
  pages = {534-552},
  title = {High-dimensional semantic space accounts of priming},
  volume = {55},
  year = {2006},
  doi = {10.1016/j.jml.2006.07.003}
}

@article{Landauer1997,
author = {Landauer, T. K. and Dumais, S. T.},
journal = {Psychological Review},
pages = {211-240},
title = {A solution to Plato's problem: The latent semantic analysis theory of acquisition, induction, and representation of knowledge},
volume = {104},
year = {1997}
}

@InProceedings{heinzerling2018bpemb,
  author = {Benjamin Heinzerling and Michael Strube},
  title = "{BPEmb: Tokenization-free Pre-trained Subword Embeddings in 275 Languages}",
  booktitle = {Proceedings of the Eleventh International Conference on Language Resources and Evaluation (LREC 2018)},
  year = {2018},
  address = {Miyazaki, Japan},
  editor = {Nicoletta Calzolari (Conference chair) and Khalid Choukri and Christopher Cieri and Thierry Declerck and Sara Goggi and Koiti Hasida and Hitoshi Isahara and Bente Maegaard and Joseph Mariani and Hélène Mazo and Asuncion Moreno and Jan Odijk and Stelios Piperidis and Takenobu Tokunaga},
  publisher = {European Language Resources Association (ELRA)},
  isbn = {979-10-95546-00-9},
  language = {english}
  }

@inproceedings{ororbia2018like,
    title = "Like a Baby: Visually Situated Neural Language Acquisition",
    author = "Ororbia, Alexander  and
      Mali, Ankur  and
      Kelly, Matthew  and
      Reitter, David",
    booktitle = "Proceedings of the 57th Annual Meeting of the Association for Computational Linguistics",
    month = jul,
    year = {2019},
    address = "Florence, Italy",
    publisher = "Association for Computational Linguistics",
    doi = "10.18653/v1/P19-1506",
    pages = "5127--5136"
}

@article{barsalou2008grounded,
author = {Barsalou, Lawrence W.},
title = {Grounded Cognition},
journal = {Annual Review of Psychology},
volume = {59},
number = {1},
pages = {617-645},
year = {2008},
doi = {10.1146/annurev.psych.59.103006.093639},
}

@article{goodman2007bayesian,
  title={A Bayesian framework for cross-situational word-learning},
  author={Goodman, Noah and Tenenbaum, Joshua and Black, Michael},
  journal={Advances in neural information processing systems},
  volume={20},
  year={2007}
}

@InProceedings{resnet50,
author = {He, Kaiming and Zhang, Xiangyu and Ren, Shaoqing and Sun, Jian},
title = {Deep Residual Learning for Image Recognition},
booktitle = {Proceedings of the IEEE Conference on Computer Vision and Pattern Recognition (CVPR)},
month = {June},
year = {2016},
doi = {10.48550/ARXIV.1512.03385},
}

@article{mali2019neural,
  title={A Neural State Pushdown Automata},
  author={Mali, Ankur Arjun and Ororbia II, Alexander G and Giles, C Lee},
  journal={IEEE Transactions on Artificial Intelligence},
  year={2021},
  publisher={IEEE}
}

@article{chung2021turing,
  title={Turing Completeness of Bounded-Precision Recurrent Neural Networks},
  author={Chung, Stephen and Siegelmann, Hava},
  journal={Advances in Neural Information Processing Systems},
  volume={34},
  year={2021}
}

@article{hahn2020theoretical,
  title={Theoretical limitations of self-attention in neural sequence models},
  author={Hahn, Michael},
  journal={Transactions of the Association for Computational Linguistics},
  volume={8},
  pages={156--171},
  year={2020},
  publisher={MIT Press}
}

@misc{stogin2020provably,
  doi = {10.48550/ARXIV.2006.03651},
  author = {Stogin, John and Mali, Ankur and Giles, C Lee},  
  title = {A provably stable neural network Turing Machine},
  publisher = {arXiv},
  year = {2020},
}

@inproceedings{merkz2022,
    title = "Seeing the advantage: visually grounding word embeddings to better capture human semantic knowledge",
    author = "Merkx, Danny  and
      Frank, Stefan  and
      Ernestus, Mirjam",
    booktitle = "Proceedings of the Workshop on Cognitive Modeling and Computational Linguistics",
    month = may,
    year = {2022},
    address = "Dublin, Ireland",
    publisher = "Association for Computational Linguistics",
    doi = "10.18653/v1/2022.cmcl-1.1",
    pages = "1--11"
}

@article{bruni2014multimodal,
  title={Multimodal distributional semantics},
  author={Bruni, Elia and Tran, Nam-Khanh and Baroni, Marco},
  journal={Journal of artificial intelligence research},
  volume={49},
  pages={1--47},
  year={2014}
}

@inproceedings{baroni2014don,
  title={Don’t count, predict! a systematic comparison of context-counting vs. context-predicting semantic vectors},
  author={Baroni, Marco and Dinu, Georgiana and Kruszewski, Germ{\'a}n},
  booktitle={Proceedings of the 52nd Annual Meeting of the Association for Computational Linguistics (Volume 1: Long Papers)},
  pages={238--247},
  year={2014}
}

@article{speer2016ensemble,
  title={An ensemble method to produce high-quality word embeddings (2016)},
  author={Speer, Robyn and Chin, Joshua},
  journal={arXiv preprint arXiv:1604.01692},
  year={2016}
}

@inproceedings{kiela2017learning,
    title = {Learning Visually Grounded Sentence Representations},
    author = {Kiela, Douwe  and Conneau, Alexis  and Jabri, Allan  and Nickel, Maximilian},
    booktitle = {Proceedings of the 2018 Conference of the North {A}merican Chapter of the Association for Computational Linguistics: Human Language Technologies, Volume 1 (Long Papers)},
    year = {2018},
    address = {New Orleans, Louisiana},
    publisher = {Association for Computational Linguistics},
    doi = {10.18653/v1/N18-1038},
    pages = {408--418}
}

@inproceedings{derby2020analysing,
  title={Analysing word representation from the input and output embeddings in neural network language models},
  author={Derby, Steven and Miller, Paul and Devereux, Barry},
  booktitle={Proceedings of the 24th Conference on Computational Natural Language Learning},
  pages={442--454},
  year={2020}
}

@article{hill2015simlex,
  title={Simlex-999: Evaluating semantic models with (genuine) similarity estimation},
  author={Hill, Felix and Reichart, Roi and Korhonen, Anna},
  journal={Computational Linguistics},
  volume={41},
  number={4},
  pages={665--695},
  year={2015},
  publisher={MIT Press One Rogers Street, Cambridge, MA 02142-1209, USA journals-info}
}

@inproceedings{finkelstein2001placing,
  title={Placing search in context: The concept revisited},
  author={Finkelstein, Lev and Gabrilovich, Evgeniy and Matias, Yossi and Rivlin, Ehud and Solan, Zach and Wolfman, Gadi and Ruppin, Eytan},
  booktitle={Proceedings of the 10th international conference on World Wide Web},
  pages={406--414},
  year={2001},
  url={https://doi.org/10.1145/371920.372094}
}

@inproceedings{agirre2009study,
    title = "A Study on Similarity and Relatedness Using Distributional and {W}ord{N}et-based Approaches",
    author = "Agirre, Eneko  and
      Alfonseca, Enrique  and
      Hall, Keith  and
      Kravalova, Jana  and
      Pa{\c{s}}ca, Marius  and
      Soroa, Aitor",
    booktitle = "Proceedings of Human Language Technologies: The 2009 Annual Conference of the North {A}merican Chapter of the Association for Computational Linguistics",
    month = jun,
    year = {2009},
    address = "Boulder, Colorado",
    publisher = "Association for Computational Linguistics",
    url = "https://aclanthology.org/N09-1003",
    pages = "19--27"
}

@inproceedings{camacho2015framework,
  title={A Framework for the Construction of Monolingual and Cross-lingual Word Similarity Datasets},
  author={Camacho-Collados, Jos{\'e} and Pilehvar, Mohammad Taher and Navigli, Roberto},
  booktitle={Proceedings of ACL (2)},
  pages={1--7},
  year={2015}
}

@article{rubenstein1965contextual,
  title={Contextual correlates of synonymy},
  author={Rubenstein, Herbert and Goodenough, John B},
  journal={Communications of the ACM},
  volume={8},
  number={10},
  pages={627--633},
  year={1965},
  publisher={ACM New York, NY, USA}
}

@inproceedings{gao2019ganbert,
  title={Code-Switching Sentence Generation by Bert and Generative Adversarial Networks.},
  author={Gao, Yingying and Feng, Junlan and Liu, Ying and Hou, Leijing and Pan, Xin and Ma, Yong},
  booktitle={INTERSPEECH},
  pages={3525--3529},
  year={2019}
}

@inproceedings{gupta2020semi,
  title={A semi-supervised approach to generate the code-mixed text using pre-trained encoder and transfer learning},
  author={Gupta, Deepak and Ekbal, Asif and Bhattacharyya, Pushpak},
  booktitle={Findings of the Association for Computational Linguistics: EMNLP 2020},
  pages={2267--2280},
  year={2020}
}

@inproceedings{lee2020balm,
  title={Modeling code-switch languages using bilingual parallel corpus},
  author={Lee, Grandee and Li, Haizhou},
  booktitle={Proceedings of the 58th Annual Meeting of the Association for Computational Linguistics},
  pages={860--870},
  year={2020}
}

@inproceedings{winata2021metaembedding,
    title = "Are Multilingual Models Effective in Code-Switching?",
    author = "Winata, Genta Indra  and
      Cahyawijaya, Samuel  and
      Liu, Zihan  and
      Lin, Zhaojiang  and
      Madotto, Andrea  and
      Fung, Pascale",
    booktitle = "Proceedings of the Fifth Workshop on Computational Approaches to Linguistic Code-Switching",
    month = jun,
    year = {2021},
    address = "Online",
    publisher = "Association for Computational Linguistics",
    doi = "10.18653/v1/2021.calcs-1.20",
    pages = "142--153"
}

@inproceedings{chi2020UGBERT,
    title = "Finding Universal Grammatical Relations in Multilingual {BERT}",
    author = "Chi, Ethan A.  and
      Hewitt, John  and
      Manning, Christopher D.",
    booktitle = "Proceedings of the 58th Annual Meeting of the Association for Computational Linguistics",
    year = {2020},
    publisher = "Association for Computational Linguistics",
    doi = "10.18653/v1/2020.acl-main.493",
    pages = "5564--5577"
}

@article{benjamini1995controlling,
  title={Controlling the false discovery rate: a practical and powerful approach to multiple testing},
  author={Benjamini, Yoav and Hochberg, Yosef},
  journal={Journal of the Royal statistical society: series B (Methodological)},
  volume={57},
  number={1},
  pages={289--300},
  year={1995},
  publisher={Wiley Online Library}
}

@inproceedings{dhar2018does,
  title={Does syntactic knowledge in multilingual language models transfer across languages?},
  author={Dhar, Prajit and Bisazza, Arianna},
  booktitle={Proceedings of the 2018 EMNLP workshop BlackboxNLP: analyzing and interpreting neural networks for NLP},
  pages={374--377},
  year={2018}
}

@article{barsalou1999perceptions,
  title={Perceptions of perceptual symbols},
  author={Barsalou, Lawrence W},
  journal={Behavioral and Brain Sciences},
  volume={22},
  number={4},
  pages={637--660},
  year={1999},
  publisher={Cambridge University Press}
}

@inproceedings{alishahi2008fast,
  title={Fast mapping in word learning: What probabilities tell us},
  author={Alishahi, Afra and Fazly, Afsaneh and Stevenson, Suzanne},
  booktitle={Proceedings of the Twelfth Conference on Computational Natural Language Learning},
  pages={57--64},
  year={2008},
  organization={Association for Computational Linguistics}
}

@article{abend2017bootstrapping,
	title = {Bootstrapping language acquisition},
	volume = {164},
	issn = {0010-0277},
	doi = {10.1016/j.cognition.2017.02.009},
	journal = {Cognition},
	author = {Abend, Omri and Kwiatkowski, Tom and Smith, Nathaniel J. and Goldwater, Sharon and Steedman, Mark},
	year = {2017},
	pages = {116 -- 143}
}

@inproceedings{frank2008bayesian, title={A {B}ayesian framework for cross-situational word-learning}, author={Frank, Michael C and Goodman, Noah D and Tenenbaum, Joshua B}, booktitle={Advances in neural information processing systems}, pages={457--464}, year={2008} }

@article{greeno1993situativity,
  title={Situativity and symbols: Response to {V}era and {S}imon},
  author={Greeno, James G and Moore, Joyce L},
  journal={Cognitive Science},
  volume={17},
  number={1},
  pages={49--59},
  year={1993},
  publisher={Wiley Online Library}
}

@inproceedings{kievit2011semantic,
  title={The semantic pictionary project},
  author={Kievit-Kylar, Brent and Jones, Michael},
  booktitle = {Proceedings of the 33rd Annual Conference of the Cognitive Science Society},
  year={2011},
  address = {Austin, TX},
  publisher = {Cognitive Science Society},
  editor={L. Carlson and C. Hoelscher and T.F. Shipley},
  pages={2229--2234},
  url={https://escholarship.org/uc/item/3zp031jv},
}

@inproceedings{xu2015show,
  title={Show, attend and tell: Neural image caption generation with visual attention},
  author={Xu, Kelvin and Ba, Jimmy and Kiros, Ryan and Cho, Kyunghyun and Courville, Aaron and Salakhudinov, Ruslan and Zemel, Rich and Bengio, Yoshua},
  booktitle={International Conference on Machine Learning},
  pages={2048--2057},
  year={2015}
}

@article{mcculloch1943logical,
  title={A logical calculus of the ideas immanent in nervous activity},
  author={McCulloch, Warren S and Pitts, Walter},
  journal={The bulletin of mathematical biophysics},
  volume={5},
  pages={115--133},
  year={1943},
  publisher={Springer}
}

@ARTICLE{Putnam2017integrated,
AUTHOR={Putnam, Michael T. and Carlson, Matthew and Reitter, David},   
TITLE={Integrated, Not Isolated: Defining Typological Proximity in an Integrated Multilingual Architecture},   
JOURNAL={Frontiers in Psychology},      
VOLUME={8},           
YEAR={2018},   	
DOI={10.3389/fpsyg.2017.02212},      
}

@InProceedings{radford2021learning,
  title = 	 {Learning Transferable Visual Models From Natural Language Supervision},
  author =       {Radford, Alec and Kim, Jong Wook and Hallacy, Chris and Ramesh, Aditya and Goh, Gabriel and Agarwal, Sandhini and Sastry, Girish and Askell, Amanda and Mishkin, Pamela and Clark, Jack and Krueger, Gretchen and Sutskever, Ilya},
  booktitle = 	 {Proceedings of the 38th International Conference on Machine Learning},
  pages = 	 {8748--8763},
  year = 	 {2021},
  editor = 	 {Meila, Marina and Zhang, Tong},
  volume = 	 {139},
  series = 	 {Proceedings of Machine Learning Research},
  publisher =    {PMLR},
  url = 	 {https://proceedings.mlr.press/v139/radford21a.html},
}

@misc{huo2021wenlan,
  doi = {10.48550/ARXIV.2103.06561},  
  author = {Huo, Yuqi and Zhang, Manli and Liu, Guangzhen and Lu, Haoyu and Gao, Yizhao and Yang, Guoxing and Wen, Jingyuan and Zhang, Heng and Xu, Baogui and Zheng, Weihao and Xi, Zongzheng and Yang, Yueqian and Hu, Anwen and Zhao, Jinming and Li, Ruichen and Zhao, Yida and Zhang, Liang and Song, Yuqing and Hong, Xin and Cui, Wanqing and Hou, Danyang and Li, Yingyan and Li, Junyi and Liu, Peiyu and Gong, Zheng and Jin, Chuhao and Sun, Yuchong and Chen, Shizhe and Lu, Zhiwu and Dou, Zhicheng and Jin, Qin and Lan, Yanyan and Zhao, Wayne Xin and Song, Ruihua and Wen, Ji-Rong},  
  title = {WenLan: Bridging Vision and Language by Large-Scale Multi-Modal Pre-Training},
  
  publisher = {arXiv},
  
  year = {2021},
}

@article{wilson2008loss,
  title={Loss of central vision and audiovisual speech perception},
  author={Wilson, Amanda and Wilson, Adam and Ten Hove, Martin W and Par{\'e}, Martin and Munhall, Kevin G},
  journal={Visual impairment research},
  volume={10},
  number={1},
  pages={23--34},
  year={2008},
  publisher={Taylor \& Francis}
}

@article{kolarik2017partial,
  title={Partial visual loss affects self-reports of hearing abilities measured using a modified version of the Speech, Spatial, and Qualities of Hearing Questionnaire},
  author={Kolarik, Andrew J and Raman, Rajiv and Moore, Brian CJ and Cirstea, Silvia and Gopalakrishnan, Sarika and Pardhan, Shahina},
  journal={Frontiers in Psychology},
  volume={8},
  pages={561},
  year={2017},
  publisher={Frontiers Media SA}
}

@inproceedings{mali2021investigating,
  title={Investigating Backpropagation Alternatives when Learning to Dynamically Count with Recurrent Neural Networks},
  author={Mali, Ankur and Ororbia, Alexander and Kifer, Daniel and Giles, Lee},
  booktitle={International Conference on Grammatical Inference},
  pages={154--175},
  year={2021},
  organization={PMLR}
}

@article{ororbia2020continual,
  title={Continual learning of recurrent neural networks by locally aligning distributed representations},
  author={Ororbia, Alexander and Mali, Ankur and Giles, C Lee and Kifer, Daniel},
  journal={IEEE Transactions on Neural Networks and Learning Systems},
  volume={31},
  number={10},
  pages={4267--4278},
  year={2020},
  publisher={IEEE}
}

@article{wang2022image,
  title={Image as a foreign language: Beit pretraining for all vision and vision-language tasks},
  author={Wang, Wenhui and Bao, Hangbo and Dong, Li and Bjorck, Johan and Peng, Zhiliang and Liu, Qiang and Aggarwal, Kriti and Mohammed, Owais Khan and Singhal, Saksham and Som, Subhojit and others},
  journal={arXiv preprint arXiv:2208.10442},
  year={2022}
}
\bibliographystyle{IEEEtran}



\end{document}